\documentclass[times,twocolumn,final]{elsarticle}

\usepackage{framed,multirow}

\usepackage{amssymb}
\usepackage{latexsym}
\usepackage{algorithm}
\usepackage{multirow}
\usepackage{graphicx}
\usepackage{booktabs}
\usepackage{caption}
\usepackage{array}

\usepackage{siunitx}
\usepackage{makecell}
\usepackage{threeparttable}
\usepackage{url}
\usepackage{xcolor}

\usepackage{pifont}

\usepackage[hidelinks]{hyperref}
\definecolor{newcolor}{rgb}{.8,.349,.1}

\usepackage[ruled,vlined,algo2e]{algorithm2e}
\SetKwComment{Comment}{$\triangleright$\ }{}

\usepackage{mathtools}
\usepackage{amsmath}

\begin{document}


\begin{frontmatter}

\title{ProDM: Synthetic Reality–driven Property-aware Progressive Diffusion Model for Coronary Calcium Motion Correction in Non-gated Chest CT}

\author[1,2]{Xinran Gong\corref{cor1}}
\ead{xgong80@gatech.edu}

\author[1]{Gorkem Durak}
\author[1]{Halil Ertugrul Aktas}
\author[1]{Vedat Cicek}
\author[1]{Jinkui Hao}
\author[1,3]{Ulas Bagci}
\author[4,5]{Nilay S. Shah}

\author[1]{Bo Zhou\corref{cor1}}
\ead{bo.zhou@northwestern.edu}

\cortext[cor1]{Corresponding author.}

\address[1]{Department of Radiology, Northwestern University, Chicago, IL, USA}
\address[2]{College of Computing, Georgia Institute of Technology, Atlanta, GA, USA}
\address[3]{Department of Biomedical Engineering, Northwestern University, Evanston, IL, USA}
\address[4]{Department of Cardiology, Northwestern University, Chicago, IL, USA}
\address[5]{Department of Preventive Medicine, Northwestern University, Chicago, IL, USA}





\begin{abstract}
Coronary artery calcium (CAC) scoring from chest CT is a well-established tool to stratify and refine clinical cardiovascular disease risk estimation. CAC quantification relies on the accurate delineation of calcified lesions, but is oftentimes affected by artifacts introduced by cardiac and respiratory motion. ECG-gated cardiac CTs substantially reduce motion artifacts, but their use in population screening and routine imaging remains limited due to gating requirements and lack of insurance coverage. Although identification of incidental CAC from non-gated chest CT is increasingly considered for it offers an accessible and widely available alternative, this modality is limited by more severe motion artifacts. We present \textbf{ProDM} (Property-aware Progressive Correction Diffusion Model), a generative diffusion framework that restores motion-free calcified lesions from non-gated CTs. ProDM introduces three key components: (1) a CAC motion simulation data engine that synthesizes realistic non-gated acquisitions with diverse motion trajectories directly from cardiac-gated CTs, enabling supervised training without paired data; (2) a property-aware learning strategy incorporating calcium-specific priors through a differentiable calcium consistency loss to preserve lesion integrity; and (3) a progressive correction scheme that reduces artifacts gradually across diffusion steps to enhance stability and calcium fidelity. Experiments on real patient datasets show that ProDM significantly improves CAC scoring accuracy, spatial lesion fidelity, and risk stratification performance compared with several baselines. A reader study on real non-gated scans further confirms that ProDM suppresses motion artifacts and improves clinical usability. These findings highlight the potential of progressive, property-aware frameworks for reliable CAC quantification from routine chest CT imaging.

\end{abstract}


\end{frontmatter}


\section{Introduction}
Cardiovascular diseases (CVDs) remain one of the most prevalent and preventable causes of death globally, with an estimation of nearly 20 million fatalities worldwide in 2022 alone ~\citep{WHO_CVDs_2025}. In the United States, CVDs are responsible for roughly every one in three mortalities ~\citep{CDC_HeartDiseaseFacts_2024}.
Detection of coronary artery calcium (CAC) can help refine CVD risk estimation, enabling timely lifestyle modification and medical intervention ~\citep{detrano2008coronary, Greenland2018CAC, Hussain2023CACAsymptomatic}. CAC score is a highly specific marker of subclinical atherosclerosis and a powerful predictor of future cardiovascular events~\citep{Greenland2018CAC, Budoff2018MESA10yr, Miedema2019YoungAdultsCAC}. Compared to other methods of evaluating coronary artery disease and CVD risk, CAC scoring acquired through chest computed tomography (CT) is a non-invasive approach and offers a reliable, reproducible, and cost-effective method for assessing coronary plaque burden and providing risk stratification of the patient developing CVDs ~\citep{Gaine2023CAC}. 

While ECG-gated cardiac CTs remain the gold standard for CAC scoring due to the ability to minimize motion artifacts, they are not routinely acquired in clinical practice due to the need for specialized imaging protocols involving ECG-gating \citep{CACPeng2023}. Additionally, many insurance do not cover such scans as it is considered a screening rather than diagnostic procedure \citep{Ali2021CACUtilityPolicy, GreenlandCalciumCTInsurance2022}. In contrast, non-gated chest CTs are far more commonly covered by insurance, and are frequently obtained for oncologic or pulmonary evaluation. Nearly twenty times more non-gated chest CTs are performed annually in the United States than gated cardiac CTs, and the ability to accurately assess incidental CAC on non-gated CT represents a major untapped opportunity for CVD prevention ~\citep{Dzaye2025CACNonCardiacCT}. In fact, studies have found that CAC scores derived from non-gated scans strongly correlated with those acquired from gated scans, and incidental findings through non-gated scans are capable of predicting major adverse cardiac events in the future ~\citep{Takx2015CAC, Almeida2020NongatedCACreliability, Yu2021IncidentalCAC, Jacobs2012LDCAC, Wetscherek2023IncidentalCAC}.  

However, accurate cardiovascular risk stratification depends on the precise representation of both the spatial distribution and intensity of calcified lesions. The Agatston score, which is the most widely clinically used metric for CAC quantification, integrates both lesion area and peak intensity within predefined thresholds to quantify calcium burden. As a result, even subtle blurring or smearing, which are common in non-gated acquisitions, can lead to substantial deviations in Agatston scores, potentially impacting clinical usability and shifting patients across risk categories \cite{Brown2000MotionArtifactCAC, BudoffAgatston2006, Taguchi2009CTArtifacts}. Consequently, despite their wide availability and potential as a substitute for gated scans, non-gated chest CTs remain limited by motion-induced degradation that undermines the accuracy of CAC quantification \citep{PARSA20241557}. Compounding this challenge is the scarcity of paired datasets, as patients rarely undergo both ECG-gated cardiac CT and non-gated chest CT \citep{Chi2021PriorNongatedCT}, making it difficult to obtain matched motion-free and motion-corrupted images for supervised learning of a motion-correction model.

Motion-compensated reconstruction (MCR) methods represent the classical efforts toward mitigating the effect of cardiac motion in CT. These techniques estimate local motion vectors fields from multi-phase or ECG-synchronized data and apply the estimated motion during reconstruction to compensate for cardiac motion ~\citep{Forthmann2008VectorFieldMotionComp, SCHIRRA2009122, Isola2010MotionCorrectedCT, TangMCR2012, Jin2018MotionCompLimitedAngle}. 
While effective in reducing motion artifacts and improving image quality, they require projection data, ECG-gating, and multiple phase reconstructions, which are not readily available and not applicable to non-gated chest CT. 

More recently, many studies began adopting deep learning approaches to learn motion correction from the image domain directly. For example, \cite{Yan2021AIReconstructionCTA} uses a VNet-based deep learning approach to perform centerline-based motion tracking and estimate the motion vector field from Coronary CT Angiography (CCTA) images. \cite{Gong2024MotionArtifactCT} proposed an attention-guided and spatial transformer–based framework that leverages cross-phase information for artifact correction, while \cite{Yao2025TemporalWeightMotionCorrection} recently introduced a temporal-weighted motion correction network with a differentiable spatial transformer module trained on simulated motion data. These works primarily target gated-cardiac CT and CCTA, which already use ECG-gating and therefore exhibit smaller motion amplitudes, whereas motion in routine non-gated chest CTs is typically more severe. Generative diffusion-based models have also demonstrated promising results for motion artifact reduction in head CT \citep{Chen2025HeadCTDiffusion} and MRI \citep{10635444, Safari2025ResMoCoDiff}, but to our best knowledge,  no study has applied such diffusion-based approaches to coronary calcium motion correction with preservation of calcified lesion properties. 

To address the aforementioned challenges and gaps in research, we propose \textbf{ProDM} (\textbf{Pro}perty-Aware Progressive Correction \textbf{D}iffusion \textbf{M}odel), a generative framework that restores motion-free coronary calcified lesions in non-gated chest CTs. Our main contributions are as follows:

 (1) \textbf{Motion Simulation Engine for Supervised Learning} We build a motion simulation data engine that generates large-scale, realistic paired datasets spanning diverse motion profiles, enabling supervised learning of motion artifact correction in non-gated chest CT scans.
 
(2) \textbf{CAC property-aware and score-driven learning.} We introduce CAC property-aware supervision that explicitly preserves calcified lesion properties through a differentiable volumetric CAC score loss, which approximates the non-differentiable Agatston scoring process.

(3) \textbf{Progressive motion correction framework for non-gated chest CT.} We developed a progressive motion correction framework that progressively removes motion artifacts and recovers the motion-free calcified lesions. This design enables stable and interpretable correction of motion artifacts in non-gated chest CTs, and is the first diffusion-based approach tailored for this task.

\section{Methods}

In this section, we detail our proposed method, including simulating realistic CAC motion artifacts (\ref{method:simulation}), the model architecture (\ref{method:arch}), implementation details (\ref{method:implementation}), and evaluation metrics (\ref{method:evaluation}).

\subsection{CAC Motion Simulation Data Engine}\label{method:simulation}

To simulate large-scale datasets of realistic CAC motion artifacts, we used the widely available cardiac-gated non-contrast CTs (i.e., calcium CT) as a starting point. Each scan was accompanied by a binary calcium mask delineating the location of calcified plaques.

We simulated 21 motion profiles to represent a range of plausible coronary calcium motion trajectories caused by cardiac or respiratory motions seen in non-gated scans, covering random jitters, translational, oscillatory, and piecewise movements. Each motion profile is represented as a trajectory-generating function $\pi(\cdot\,;\phi)$, where $\pi$ specifies the functional form of the motion (e.g., sinusoidal, linear displacement, etc.), and $\phi$ represents the corresponding profile-specific parameters such as directions in 3D, displacement magnitude, velocity, and temporal phase, where applicable. Together, $(\pi, \phi)$ defines the complete motion trajectory used to identify the position of the calcified lesion at each partial angle of the forward projection. 
We intentionally modeled displacement vectors primarily in the axial plane (xy-direction), where motion magnitude is typically largest, while the z-direction displacement was kept relatively small to reflect realistic motion patterns seen in non-gated chest CTs. For each simulation, the number of projection angles $N$ was randomly sampled from $\{180, 360, 540, 720, 1080\}$ to simulate different temporal resolutions and acquisition durations, producing motion artifacts of varying levels of severity and trajectory smoothness. Randomized perturbations were added to the underlying parameters $\phi$ to introduce stochastic variability across samples, ensuring that the coronary calcium motion artifacts exhibited per-patient differences.

\begin{figure*}[!t]
    \centering{  
    \includegraphics[width=1\linewidth]{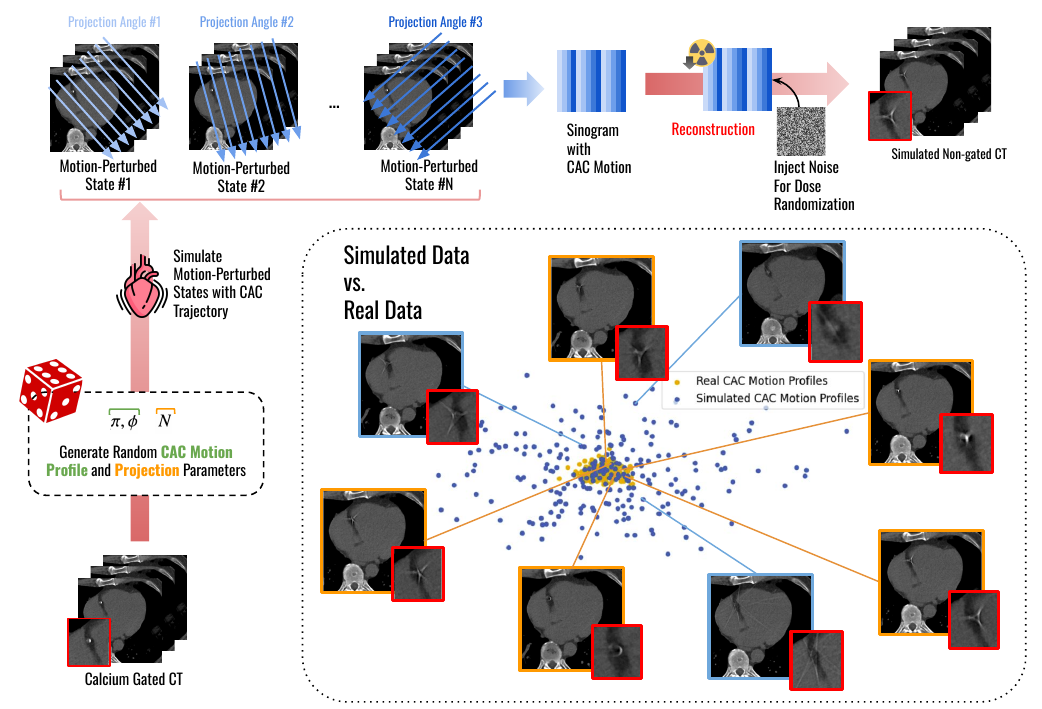}
    }
    \caption{{Overview of CAC Motion Data Engine for simulating non-gated CT from gated calcium CT. The displacement trajectory of the calcified lesion is computed based on the selected CAC motion profile and projection parameters with added stochasticity, which is applied to the motion-free gated calcium CT introducing motion perturbation to the calcified lesion at each partial angle of the forward projection. The resulting sinogram then undergoes inverse Radon transform to obtain the simulated non-gated CT. The bottom-right panel demonstrates that the simulated CAC motion profiles cover a superset of real motion patterns, including simulated scans with motion profiles similar to real CAC motion (orange) as well as additional, less realistic scans (blue) introduced to increase diversity.}}
    \label{fig-motion-simulation-pipeline}
\end{figure*}

The overall motion simulation process, which includes how $(\pi, \phi)$ defines the displacement trajectory, is outlined in high-level in Algorithm~\ref{alg:motion-simulation} and shown in Figure \ref{fig-motion-simulation-pipeline}. To briefly summarize, calcium regions were first removed via inpainting and reinserted at new positions at each projection angle determined by the chosen motion trajectory. Forward projections were computed using the Radon transform to form motion-perturbed sinograms for each projection angle, followed by reconstruction using the inverse Radon transform to obtain synthetic motion-corrupted CT volumes. Further details of the simulated motion profiles, the corresponding parameters, as well as the value ranges of these parameters are available in \textbf{Appendix} ~\ref{tab:motion_profiles}.

\begin{algorithm}[!t]
\caption{Motion Modeling Framework for Simulating Coronary Calcium Motion}
\label{alg:motion-simulation}
\KwIn{ECG-gated calcium CT volume $x_0$; binary calcium mask $m$.}
\KwOut{Synthetic non-gated motion-perturbed reconstruction $y$.}

\textbf{1. Inpaint to remove calcium:}\\
$x_0' \leftarrow \mathrm{Inpaint}(x_0, m)$\\

\textbf{2. Select motion profile and parameters:}\\
choose $\pi$ with params $\phi$\\

\textbf{3. Choose projection count:}\\
$N \in \{180, 360, 540, 720, 1080\}$; set angles $\Theta=\{\theta_i\}_{i=1}^N$\\

\textbf{4. Generate trajectory:}\\
compute displacements $\{\Delta x(\theta_i)\}_{i=1}^N$ from $(\pi,\phi,N)$\\

\textbf{5. Angle-wise forward projection:}\\
\For{$i=1$ \KwTo $N$}{
  Shift calcium by $\Delta x(\theta_i)$ using $m$ to form instantaneous image $x_i$ at partial angle $\theta_i$ (reinsert original calcium values into $x_0'$). \\
  $p_i \leftarrow \mathrm{Radon}(x_i,\theta_i)$ is the partial angle projection at angle $i$\\ 
}
\textbf{6. Stack to sinogram:} $S \leftarrow [\,p_1,\ldots,p_N\,]$\\

\textbf{7. Reconstruct:} $y \leftarrow \mathrm{iRadon}(S,\Theta)$

\end{algorithm}

\subsection{Progressive Property-Aware Motion Correction Model} \label{method:arch}

Our proposed framework adopts a diffusion model as a backbone with a direct bridge between the motion-corrupted and motion-free patches. This formulation is suitable for our motion correction task because it constrains the diffusion process to begin and end with meaningful images rather than arbitrary noise, allowing the model to learn a more stable, interpretable, and structured denoising trajectory that progressively transforms motion-corrupted inputs toward their motion-free counterparts. We embed calcium-specific priors through a calcium consistency loss into the diffusion process, so that the recovered motion-free calcified lesion preserves the original intensity, quantity, and spatial distribution. 

\begin{figure*}[!t]
    \centering{  
    \includegraphics[width=0.95\linewidth]{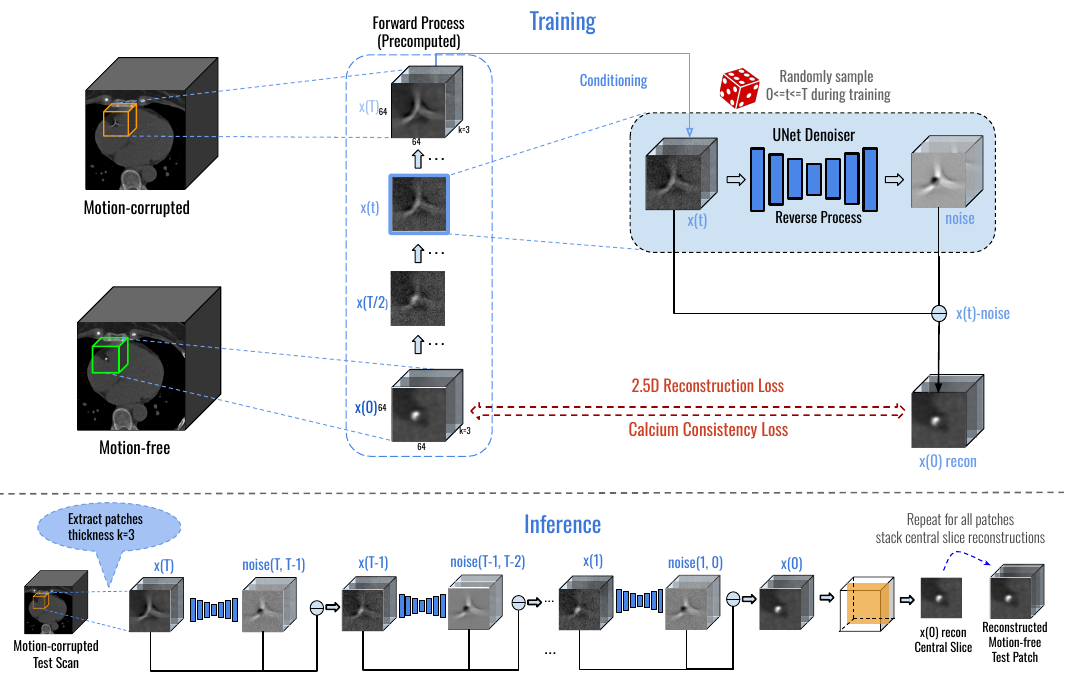}
    }
    \caption{{Overview of the ProDM framework demonstrating the training and inference processes for CAC motion correction in non-gated chest CTs. The top figure shows training pipeline, where motion-corrupted non-gated CT volumes (orange) and corresponding motion-free gated calcium CT volumes (green) are used to construct supervised training pairs. Training is set up through a precomputed forward diffusion process where noise is progressively added to obtain intermediate states $x(t)$, a UNet denoiser learns to predict this noise at randomly sampled timesteps, guided by reconstruction loss and calcium consistency loss. Bottom figure shows the sliding-window inference pipeline used at test time, where all motion-corrupted patches within an ROI undergo progressive motion correction. The central slices of neighboring context windows are stacked together to form the final motion-corrected reconstruction of the ROI.
    }}
    \label{fig-model-arch}
\end{figure*}

\subsubsection{Forward Diffusion Process}
Let $\{x_t\}_{t=0}^T$ denote a diffusion trajectory between $x_0$ and $y$ where $T$ is the total number of timesteps to transition from the motion-free to the motion-corrupted image with gradual and controlled addition of noise. The forward process is defined by iteratively interpolating states between the two deterministic endpoints:
\begin{equation}
\label{eq:bbdm-forward}
q(x_t \mid x_0, y)
= \mathcal{N}\!\left(
x_t;\,
(1 - \alpha_t)\, x_0 + \alpha_t\, y,\;
\delta_t \mathbf{I}
\right),
\end{equation}
where $\alpha_t = \frac{t}{T}\in [0,1]$ is a monotonically increasing schedule that controls the interpolation between $y$ and $x_0$, and $\delta_t=2(\alpha_t - \alpha_t^2)$ is the variance of noise added at each step $t$. At $t = 0$, we have the motion-free image $x_0$; at $t = T$, the distribution of the image centers at $y$. This approach ensures that all intermediate states remain visually plausible CT images, rather than potentially drifting away into noisy, unrealistic states. Each intermediate state $x_t$ can be represented as:
\begin{equation}
\label{eq:xt-reparam}
x_t \;=\; (1 - \alpha_t)\, x_0 \;+\; \alpha_t\, y \;+\; \sqrt{\delta_t}\, \varepsilon,
\qquad \varepsilon \sim \mathcal{N}(0, \mathbf{I}).
\end{equation}

\subsubsection{Reverse Denoising Process}
The reverse process learns to recover the motion-free image from noisy, motion-corrupted image by iteratively estimating $x_{t-1}$ from $x_t$. Same as standard diffusion frameworks, the reverse process follows a Markov process defined by:
\begin{equation}
\label{eq:bbdm-reverse}
p_\theta(x_{t-1} \mid x_t, y)
= \mathcal{N}\!\left(
x_{t-1};\,
\mu_\theta(x_t, y, t),\;
\tilde{\delta_t} \mathbf{I}
\right),
\end{equation}
where $\mu_\theta$ is the predicted mean value of the noise and $\tilde{\delta_t}$ is the variance of noise at step $t$, obtained in closed form from the forward variance schedule \citep{li2023bbdm}. The reparameterization technique in DDPM ~\citep{DDPM2020} was adopted and $\mu_\theta$ can be reformulated as follows:
\begin{equation}
\mu_\theta(x_t, y, t)
=
c_{x,t}\, x_t
+
c_{y,t}\, y
+
c_{e,t}\,\epsilon_\theta(x_t,y,t),
\end{equation}
where
\begin{equation}
\begin{aligned}
c_{x,t} &= 
\frac{\delta_{t-1}}{\delta_t}
\frac{1 - \alpha_t}{1 - \alpha_{t-1}}
+
\frac{\delta_{t|t-1}}{\delta_t}
(1 - \alpha_{t-1}), \\[8pt]
c_{y,t} &= 
\alpha_t - \alpha_{t-1}
\frac{1 - \alpha_t}{1 - \alpha_{t-1}}
\frac{\delta_{t-1}}{\delta_t}, \\[8pt]
c_{e,t} &= 
(1 - \alpha_{t-1})
\frac{\delta_{t|t-1}}{\delta_t}, \\[8pt]
\epsilon_\theta(x_t,y,t) &= \big(\alpha_t (y - x_0) + \sqrt{\delta_t}\, \varepsilon\big).
\end{aligned}
\end{equation}
Note that $c_{x,t}=$, $c_{y,t}$, and $c_{e,t}$ are non-trainable, time-dependent coefficients determined by the diffusion schedule. Instead of directly predicting $\mu_\theta(x_t, y, t)$, we can train a neural network for learning to predict the noise term $\epsilon_\theta(x_t,y,t)$ at each timestep, which can be used to recover the reconstructed motion-free image at inference time. To maintain consistency with the motion correction setting, the corrupted image $y$ is provided as a conditioning input throughout the denoising path.

With this formulation, the training objective is given by optimizing the Evidence Lower-Bound (ELBO) which can be simplified as 
\begin{equation}
\label{eq:main-loss-elbo}
ELBO
=
\mathbb{E}_{x_0, y, \varepsilon}
\left[
\left\|
\alpha_t(y - x_0) + \sqrt{\delta_t}\epsilon - \epsilon_\theta(x_t, y, t)
\right\|_2^2
\right].
\end{equation}

\subsubsection{Calcium Property-Aware Consistency Learning}
CAC scoring methods, such as the Agatston and Volume scores ~\citep{gupta2022coronary}, are highly sensitive to small intensity or structural changes in calcified regions. Therefore, even subtle motion artifacts can result in large deviations in computed CAC scores, which in turn affect downstream CVD risk stratification. Due to the relatively small size of calcified lesions in proportion to the context image, this is difficult to enforce using diffusion alone. In addition to learning to restore overall image fidelity through progressive motion correction, our model explicitly enforces a calcium-specific consistency loss to preserve the quantitative accuracy and spatial distribution of the calcium regions. 

Since the Agatston score is the most widely adopted CAC scoring method in clinical practice, it would ideally be incorporated directly into the training objective to enforce calcified lesion consistency. However, because the Agatston score depends on the maximum voxel intensity within each calcified lesion, it is inherently non-differentiable and therefore unsuitable for gradient-based optimization. To address this, we employ the Volume score \citep{Thomas2018CACVolumeScore} as a differentiable surrogate that approximates the Agatston score by summing the intensities of all voxels exceeding the clinical threshold of 130 Hounsfield Unit (HU). 

For any voxel with intensity $x_i$, the soft mask function is defined as:
\begin{equation}
\operatorname{SoftMask}(x_i) = 
\sigma\!\left(\frac{x_i - 130}{\tau}\right),
\label{eq:softmask}
\end{equation}
where $\sigma(\cdot)$ denotes the sigmoid function and $\tau$ controls the softness of the 130 HU clinical threshold for coronary calcium.

Given motion-free reference $x_0$ and reconstructed motion-free sample $\hat{x_0}$, the resulting calcium score consistency loss is defined as:


\begin{equation}
\mathcal{L}_{\text{vol}} =
\mathbb{E}\!\left[
\mathrm{MSE}\!\left(
\log(1 + S_{\text{vol}}(\hat{x}_0)),
\log(1 + S_{\text{vol}}(x_0))
\right)
\right]
\end{equation}

where $S_\text{vol}(x)$ computes the differentiable volume score as
\begin{equation}
S_\text{vol}(x) = v_\text{voxel} \sum_{i\in V} \operatorname{SoftMask}(x_i),
\label{eq:vol-score}
\end{equation}

This loss encourages the reconstructed motion-free image $\hat{x}_0$ to yield a calcium burden consistent with the reference $x_0$.

Training and sampling algorithms are described in Algorithms \ref{alg-training} and \ref{alg-inference}, respectively. 

\begin{algorithm}[!t]
\caption{{Training Scheme of ProDM}}
\KwIn{
Motion-free batch $\mathcal{B}_0 = \{x_0\}$
\tcp*{$x_0$: motion-free CAC patches}

Motion-corrupted batch $\mathcal{B}_T = \{y\}$
\tcp*{$y$: motion-corrupted patches paired with $x_0$}

\textbf{Other inputs:} total diffusion steps $T$, denoiser $\epsilon_\theta$, schedules $\{\alpha_t\}_{t=1}^T$, $\{\delta_t\}_{t=1}^T$, calcium consistency loss weight $\lambda$\;
}
\KwOut{Total loss $\mathcal{L}_{\text{total}}$ for one optimization step}

\textbf{Sample diffusion step and noise:} \\
$t \sim \mathcal{U}\{1, \dots, T\}$ \\
$\epsilon \sim \mathcal{N}(0, I)$ \\

\textbf{Construct noisy intermediate state and target noise:} \\
$x_t \leftarrow (1 - \alpha_t) x_0 + \alpha_t y + \sqrt{\delta_t}\,\epsilon$ \\
$n_t \leftarrow \alpha_t (y - x_0) + \sqrt{\delta_t}\,\epsilon$ \\

\textbf{Forward pass through the denoiser:} \\
$\hat{\epsilon}_t \leftarrow \epsilon_\theta(x_t, y, t)$ \\
$\hat{x_{0,t}} \leftarrow x_t - \hat{\epsilon}_t$ \\

\textbf{Compute auxiliary loss:} \\
$\mathcal{L}_{\text{calc}} \leftarrow \text{CalciumConsistencyLoss}(x_0, \hat{x_{0,t}})$ \\

\textbf{Aggregate total loss:} \\
$\mathcal{L}_{\text{total}} \leftarrow \|\;n_t - \hat{\epsilon}_t\;\|_2^2 + \lambda\mathcal{L}_{\text{calc}}$ \\

\textbf{Update denoiser parameters (Adam):} \\
$\theta \leftarrow \text{AdamUpdate}(\theta, \nabla_\theta \mathcal{L}_{\text{total}}, \eta, \beta_1, \beta_2)$ \\
\quad where $\eta$ is the learning rate, and $\beta_1, \beta_2$ are Adam momentum coefficients \\

\textbf{End on convergence}\\
\label{alg-training}
\end{algorithm}

\begin{algorithm}[!t]
\caption{{ProDM Inference Process}}
\KwIn{
Motion-corrupted ROI $y = x_T$\;

Context window size $(H, W, k)$ and extraction stride\;

Total diffusion steps $T$, trained denoiser $\epsilon_\theta$\;
}
\KwOut{Reconstructed motion-suppressed ROI $\hat{x}_0$}

\textbf{Extract context windows:} \\
$\{x_T^{(i)}\}_{i=1}^N \leftarrow \text{ExtractWindows}(y, H, W, k)$ \\
\quad \tcp*{$x_T^{(i)}$: 2.5D context window centered at slice $i$}

\textbf{Sliding window inference:} \\
Initialize list $\text{recon\_roi} \leftarrow [\ ]$ \\

\For{$i = 1$ \KwTo $N$}{
    $x_T \leftarrow x_T^{(i)}$ \\[2pt]
    
    \textbf{Reverse diffusion process:} \\
    \For{$t = T, T-1, \dots, 1$}{
        $\hat{n}_{t \rightarrow t-1} \leftarrow \epsilon_\theta(x_t, y, t)$ \\
        $x_{t-1} \leftarrow x_t - \hat{n}_{t \rightarrow t-1}$ \\
    }
    
    \textbf{Extract central slice:} \\
    $x_0^{(i)} \leftarrow \text{CentralSlice}(x_0)$ \\
    Append $x_0^{(i)}$ to $\text{recon\_roi}$ \\
}

\textbf{Stack slice-wise reconstructions:} \\
$\hat{x}_0 \leftarrow \text{Stack}(\text{recon\_roi})$ \\

\Return $\hat{x}_0$
\label{alg-inference}
\end{algorithm}

\subsection{Implementation Details}
\label{method:implementation}

\paragraph{\textbf{Data Preprocessing Implementation Details}}
Before training, both motion-free and simulated motion-corrupted CT scans are preprocessed and converted into 3D patches. Each CT volume was intensity-clipped to the range $[-200, 800]$ Hounsfield Units (HU) to focus on the coronary calcium, and then normalized to $[0, 1]$. Because coronary calcium occupies only a small portion of the CT volume, directly performing uniform patch extraction over the entire axial plane would result in a large imbalance between background and calcium-containing regions. To address this, we first identified regions of interest (ROIs) that contained either coronary calcium along with adjacent background tissue, or background-only regions, based on the provided calcium masks. This approach ensures that the model learns both the local calcium morphology and the surrounding anatomical context necessary for accurate motion correction. Each ROI was extracted as a 3D block of size $(64, 64, 16)$.

We used 435 gated calcium CTs publicly available from the Stanford COCA dataset. For each scan, we simulated 21 motion-corrupted non-gated chest CT scans, with a total of 9135 scans. We adopt an 80\%-20\% split between training and testing, which corresponds to 7228 training scans and 1847 testing scans, with no overlap of scans between splits. From the training scans, we extracted a total of 269,206 patches. The relatively large number of training patches arises from dense patch extraction around calcified lesions with small random spatial offsets, serving as a form of data augmentation to improve robustness against spatial translation, even though the number of unique calcified lesions is substantially smaller. From the testing scans, we extracted 13,813 patches for evaluation. We intentionally extracted fewer patches during testing for two reasons. Firstly, many patches obtained through dense patch extraction correspond to the same calcified lesion with minor spatial shifts and therefore provide redundant information for evaluation. Secondly, diffusion-based inference is computationally expensive, making exhaustive inference on all possible patches impractical without providing useful additional insight. The selected testing patches were used consistently across all methods to ensure fair comparison.

\paragraph{\textbf{Framework Implementation Details}}
Our framework was implemented in \texttt{PyTorch} and trained on a single NVIDIA RTX~6000~Ada GPU. We used the Adam optimizer with a learning rate of $2\times10^{-4}$, $\beta_1 = 0.9$, $\beta_2 = 0.999$, and a weight decay of $1\times10^{-4}$.
The batch size was set to $64$ for all experiments. For efficiency, we adopt a 2.5D training strategy, similar to recent multi-view diffusion approaches that leverage orthogonal slice ensembles to approximate full 3D generation while reducing computational cost~\citep{Chen202425DMA}. Training is performed using adjacent, overlapping patches of size ($64, 64, 3$) along the axial (z) direction extracted from the ROI, and inference is conducted using a sliding-window strategy to aggregate the central slices of the predicted motion-corrected patches to recover the reconstructed motion-corrected volume. The calcium consistency loss weight was empirically set to $\lambda = 20$. For computing the soft calcium region mask, we used a softness parameter of $\tau = 60$. The diffusion process uses $T = 1000$ total timesteps, and the sampling interval is set to $100$ for efficient inference. The denoising network $\epsilon_{\theta}$ follows a UNet architecture adapted from the OpenAI diffusion implementation (\texttt{UNetModel}). The network takes 6 input channels, corresponding to a motion-corrupted patch and a context patch (each with depth $3$), and predicts $3$ output channels. It has a base channel width of $128$ and uses channel multipliers $[1,\,2,\,2,\,2]$ across levels, attention is applied at spatial resolutions \{$16$,$8$\} with a channel width of $64$ per attention head. Temporal embeddings are used to encode the diffusion timesteps.

\subsection{Evaluation Strategies and Metrics}
\label{method:evaluation}

 We evaluate the performance of ProDM using both quantitative and qualitative assessments. For quantitative evaluation on synthetic data, we compare ProDM with several commonly used image-to-image translation baselines, including UNet \citep{ronneberger2015unet}, cGAN \citep{mirza2014conditionalGAN}, DDPM \citep{DDPM2020}, and the BBDM \citep{li2023bbdm}. Because the primary objective of motion correction is to improve the reliability of CAC quantification, we measure performance using four calcified lesion-related metrics: (1) \textbf{Mean absolute error (MAE) of Agatston scores}: to measure how well the corrected results preserve the calcium burden compared to ground truth; (2) \textbf{CAC severity grade accuracy}: to evaluate downstream CVD risk stratification according to clinical guidelines \citep{Shreya2021CACIndicator} by comparing predicted severity categories with ground truth; (3) \textbf{Dice loss}: to assess the preservation of calcified lesion spatial distribution after motion correction compared to ground truth; (4) \textbf{Pearson correlation of Agatston scores}: to assess overall agreement between predicted and ground truth Agatston scores.

To further evaluate the generalizability of our method, we conducted a reader study with two experienced radiologists using real non-gated chest CT scans from the Stanford COCA dataset. We extracted 30 motion-corrupted patches containing coronary calcium and generated corresponding motion-corrected reconstructions using both the UNet baseline and ProDM. For each de-identified sample, readers rated motion suppression, clinical usability, and vessel fidelity on a scale from 1~to~5, where 1 represents the lowest quality and 5 represents the highest. Further details of the evaluation rubric are available in \ref{Appendix-A} Table \ref{tab:likert_scale}. Readers were also asked to provide an overall ranking of the three methods (original, UNet, and ProDM). To ensure a fully blinded assessment, the three method outputs for each sample were randomly ordered and labeled as M1, M2, and M3 without revealing the underlying method. In addition, the ordering of samples themselves was randomized across the study to prevent potential recall or ordering bias.

\section{Results}

\subsection{Comparison with Baseline Methods}
In this section, we compare the proposed ProDM framework on the simulated motion test dataset against several models commonly used in image-to-image translation tasks for medical imaging.

\begin{figure*}[!h]
    \centering  
    \includegraphics[width=0.9\linewidth]{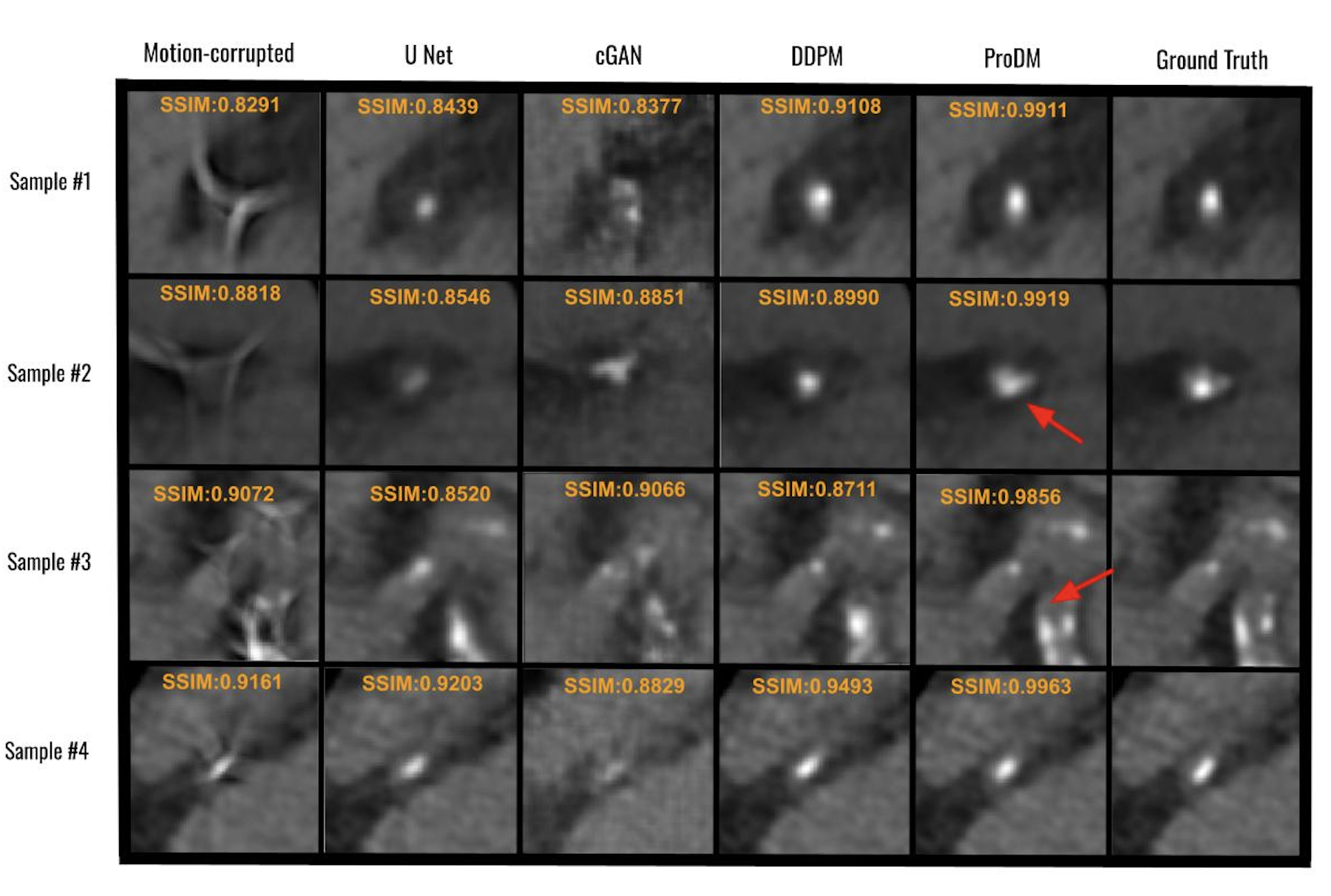}
    \vspace{-5pt}
    \caption{
    Qualitative comparison of motion correction performance of different models on simulated motion-corrupted samples. SSIM with the ground truth is shown for reference.
    }
    \vspace{-10pt}
    \label{fig:visualize-performance-comparison}
\end{figure*}

Figure~\ref{fig:visualize-performance-comparison} presents qualitative comparisons of the evaluated methods on representative motion-corrupted test samples with varying artifact types and severity. For easier assessment of qualitative similarity with the ground truth, the Structural Similarity Index (SSIM) between each reconstructed motion-suppressed image and the ground truth is provided. The UNet produced slightly higher SSIM for two samples but lower for the other two, showing clear over-smoothing and poor calcium region intensity preservation. cGAN achieved marginally higher SSIM than the uncorrected case but failed to maintain the correct spatial distribution and calcium intensity, often producing under-represented lesions. DDPM generated visually sharper results and achieved higher SSIM while improving intensity consistency; however, the spatial distribution of calcified lesions were evidently distorted in samples \#2 and \#3. In contrast, ProDM achieved the highest SSIM across all samples ($0.9911$, $0.9919$, $0.9856$, $0.9963$), closely resembling the motion-free reference. The reconstructed calcium regions retained accurate shapes and intensity patterns, with only slight underestimation of intensity observed at the lesion center in sample~\#2. It is worth noting that a higher SSIM does not always imply superior motion correction quality, as illustrated by cGAN, which achieved moderate SSIM yet failed to restore the true spatial and intensity distribution of calcified lesions.

Table~\ref{tab:performance-comparison} shows the quantitative performance comparison across different methods and evaluation metrics. It is evident that without motion correction, Agatston score–related properties are poorly preserved in the motion-corrupted samples, as indicated by a high Agatston Score MAE of $37.876\pm0.588$, low Agatston grade accuracy of $42.614\pm0.544\%$, a high average Dice loss of $0.853\pm0.002$. Pearson correlation between predicted and ground truth scores is $0.953$. The UNet model achieved moderate improvement in Agatston grade accuracy and Dice loss but exhibited higher Agatston Score MAE and lower Pearson correlation. cGAN performed the worst among all models, with the highest Agatston MAE, lowest grade accuracy, and weakest correlation, despite a marginal reduction in Dice loss compared with uncorrected inputs. DDPM demonstrated better performance, achieving lower MAE, higher grade accuracy, and substantially lower Dice loss, though it showed slightly reduced Pearson correlation. BBDM significantly improved motion correction across all metrics but still struggled to maintain accurate calcified lesion distribution and grade accuracy in certain cases. 

\begin{table*}[htbp!]
\centering
\small
\begin{threeparttable}
\caption{Motion correction performance comparison on synthetic motion-corrupted patches derived from Stanford COCA.}
\label{tab:performance-comparison}
\setlength{\tabcolsep}{7pt}
\begin{tabular}{
  l
  S[table-format=3.1(1)]   
  S[table-format=2.1(1)]   
  S[table-format=1.3(3)]   
  S[table-format=1.3(3)]   
}
\toprule
\textbf{Model} & {\textbf{Agatston MAE} $\downarrow$} & {\textbf{Grade Accuracy}(\%) $\uparrow$} & {\textbf{Avg. Dice Loss} $\downarrow$} & {\textbf{Agatston Pearson Corr.} $\uparrow$}\\
\midrule
No correction               & {$37.876\pm 0.588$} & {$42.614 \pm 0.544$} & {$0.853 \pm 0.002$}& {$0.953$} \\
\midrule
UNet                     & {$45.163 \pm 0.829$} & {$68.146 \pm 0.512$} & {$0.403 \pm 0.002$} & {$0.929$}\\
cGAN                     & {$59.729 \pm 8.256$} & {$35.431 \pm 2.309$} & {$0.806 \pm 0.011$} & {$0.748$}\\
DDPM      & {$33.659 \pm 0.985$} & {$82.571 \pm 0.530$} & {$0.210 \pm 0.002$} & {$0.942$}\\
BBDM                         & {$10.644\pm0.261$} & {$92.735 \pm 0.285$} & {$0.131 \pm 0.001$} & {$0.992$}\\
ProDM (Ours)      & \textbf{9.391 $\pm$ 0.257} & \textbf{95.334 $\pm$ 0.232} & \textbf{0.078 $\pm$ 0.001} & \textbf{0.993}\\
\bottomrule
\end{tabular}

\begin{tablenotes}[flushleft]
\footnotesize
\item \textbf{Agatston severity categories.} 0 (No calcification), 1-10 (Minimal), 11-100 (Mild), 101-400 (Moderate), $>400$ (Severe). Grade accuracy = proportion of cases where predicted grade matches ground-truth grade.
\item \textbf{Dice loss.} $1 - \text{Dice coefficient}$ between predicted calcium mask and ground-truth calcium region on the motion-free (gated) reference.
\end{tablenotes}
\end{threeparttable}
\end{table*}

Among all models, ProDM achieved the best overall performance across all metrics. After motion correction with ProDM, the Agatston Score MAE decreased to \textbf{9.391$\pm$0.257}, the grade accuracy increased to \textbf{95.334$\pm$0.232\%}, the Dice loss between predicted and ground-truth calcium regions dropped to \textbf{0.078$\pm$0.001}, and the Pearson correlation improved to \textbf{0.993}, demonstrating its superiority in preserving CAC score accuracy, calcified lesion spatial and intensity distributions, as well as scoring consistency.

\begin{figure*}[!h]
    \centering  
    \includegraphics[width=1\linewidth]{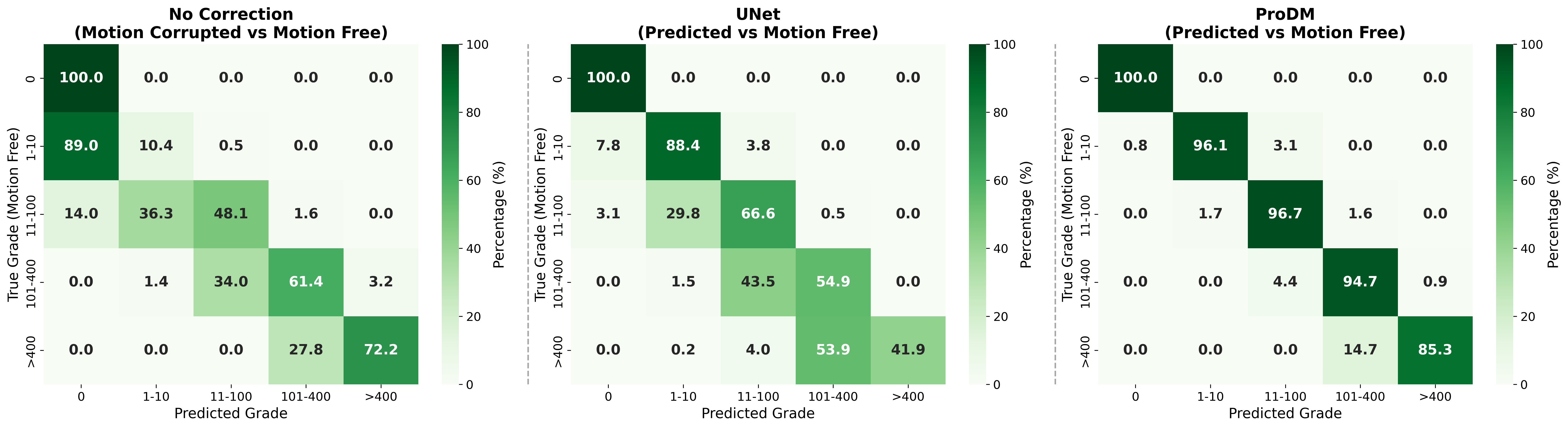}
    \vspace{-5pt}
    \caption{
    Percentage-based confusion matrices for Agatston score–based cardiovascular risk stratification using three methods: no correction (motion-corrupted), UNet baseline, and ProDM. Each cell represents the proportion of subjects within a given ground-truth risk category (rows) assigned to a predicted category (columns). Subjects are grouped into five Agatston-based risk levels: 0 (no CAC), 1–10 (minimal CAC), 11–100 (mild CAC), 101–400 (moderate CAC), and $>$400 (severe CAC).
    }
    \vspace{-10pt}
    \label{fig-confusion-matrices}
\end{figure*}

We also assess the model's performance on per-class severity category accuracy for risk stratification. Figure ~\ref{fig-confusion-matrices} shows the percentage-based confusion matrices of Agatston score cardiovascular risk categories of the uncorrected baseline, UNet-corrected results, and ProDM-corrected results. Without motion correction, the risk stratification accuracy is poor, especially in the minimal CAC score range, where only $10.4\%$ cases are correctly categorized and an overwhelming $89.0\%$ of cases are incorrectly categorized as no CAC, and the accuracy for the mild CAC score range is also quite low at $48.1\%$. This indicates that without motion correction, more subtle and lower intensity calcified lesions are most likely to be negatively impacted in terms of CAC score accuracy. UNet-corrected outputs exhibited a lot better severity category accuracy in the minimal ($88.4\%$) to mild ($66.6\%$) CAC score ranges, but the performance for moderate ($61.4\% \rightarrow 54.9\%$) to severe ($72.2\% \rightarrow 41.9\%$) score ranges decreased significantly compared to uncorrected samples. This is consistent with the observation that UNet has the tendency to oversmooth and result in decreased intensities in predictions than the ground truth. ProDM performed the best and showed significant improvement compared to the other two methods, with excellent per-category accuracy across the minimal (\textbf{96.1\%}), mild (\textbf{96.7\%}), moderate (\textbf{94.7\%}), and severe (\textbf{85.3\%}) score ranges.

\subsection{Ablation Studies}
In order to better understand how ProDM performs under different configurations, we conduct extensive ablation studies on assessing the role of calcium consistency loss and axial context thickness. 
\subsubsection{Effect of Calcium Consistency Loss on Risk Stratification}
We first assess the role of calcium consistency loss in ProDM by comparing the performance with and without this auxiliary loss, both quantitatively and qualitatively. For quantitative comparison, \textbf{Table} \ref{tab:ablative-performance-comparison} shows the per-class severity category accuracy, precision, recall, and F1 score of the motion correction methods. Without motion correction, the performance is poor as expected, particularly for the 1–10 (minimal) score range, where precision, recall, and F1 are 0.150, 0.104, and 0.123, respectively.
ProDM without the calcium consistency loss improves performance across all severity categories. However, adding the calcium consistency loss provides significant further improvement in almost all metrics across all severity categories. The most notable improvements occur in the minimal (1–10) and mild (11–100) score ranges, where recovering small calcified lesions is most challenging. For example, precision and recall in the 1–10 range increase from 0.932 and 0.914 (no-loss) to 0.972 and 0.961 (with-loss), yielding an F1 score of 0.966. Similar trends are seen in higher score ranges as well. One exception is the moderate range (101-400), which sees a slight decrease in precision (0.940$\rightarrow$0.936) when calcium consistency loss is added, but this is accompanied by an improvement in recall (0.929$\rightarrow$0.947). As a result, the overall F1 score still increases (0.934$\rightarrow$0.941).

\textbf{Figure} \ref{fig:ablative-calc-consistency-loss} shows the qualitative comparison of ProDM without vs with calcium consistency loss on some sample cases compared against the ground truth. SSIM scores against ground truth are displayed for visual reference. With calcium consistency loss, the SSIM scores are consistently higher than the same cases without calcium consistency loss. Samples \#2 and \#3 show the most evident qualitative improvement. For these cases, without calcium consistency loss, ProDM did not successfully recover the correct calcified lesion shape, which was recovered successfully with its addition, further highlighting the significance of the calcium consistency loss. 

\begin{table*}[t]
\small
\centering
\caption{Per severity grade performance comparison across score ranges for uncorrected baseline, ProDM without calcium consistency loss, and ProDM with calcium consistency loss.}
\label{tab:ablative-performance-comparison}
\begin{tabular}{lcccccccccccc}
\toprule
& \multicolumn{3}{c}{1-10 (Minimal)} 
& \multicolumn{3}{c}{11-100 (Mild)} 
& \multicolumn{3}{c}{101-400 (Moderate)}
& \multicolumn{3}{c}{$>$400 (Severe)} \\
\cmidrule(lr){2-4} \cmidrule(lr){5-7} \cmidrule(lr){8-10} \cmidrule(lr){11-13}
Method of Motion Correction
& Prec & Rec & F1
& Prec & Rec & F1
& Prec & Rec & F1
& Prec & Rec & F1 \\
\midrule
Uncorrected
& 0.150 & 0.104 & 0.123
& 0.711 & 0.481 & 0.574
& 0.863 & 0.614 & 0.717
& 0.847 & 0.722 & 0.780 \\
No Calc. Consistency Loss
& 0.932 & 0.914 & 0.923
& 0.953 & 0.945 & 0.949
& \textbf{0.940} & 0.929 & 0.934
& 0.957 & 0.853 & 0.902 \\
With Calc. Consistency Loss
& \textbf{0.972} & \textbf{0.961} & \textbf{0.966}
& \textbf{0.956} & \textbf{0.967} & \textbf{0.961}
& 0.936 & \textbf{0.947} & \textbf{0.941}
& \textbf{0.959} & \textbf{0.853} & \textbf{0.903} \\
\bottomrule
\end{tabular}
\end{table*}

\subsubsection{Impact of Calcium Consistency Loss Weight}
\begin{figure*}[!t]
    \centering
    \includegraphics[width=0.85\linewidth]{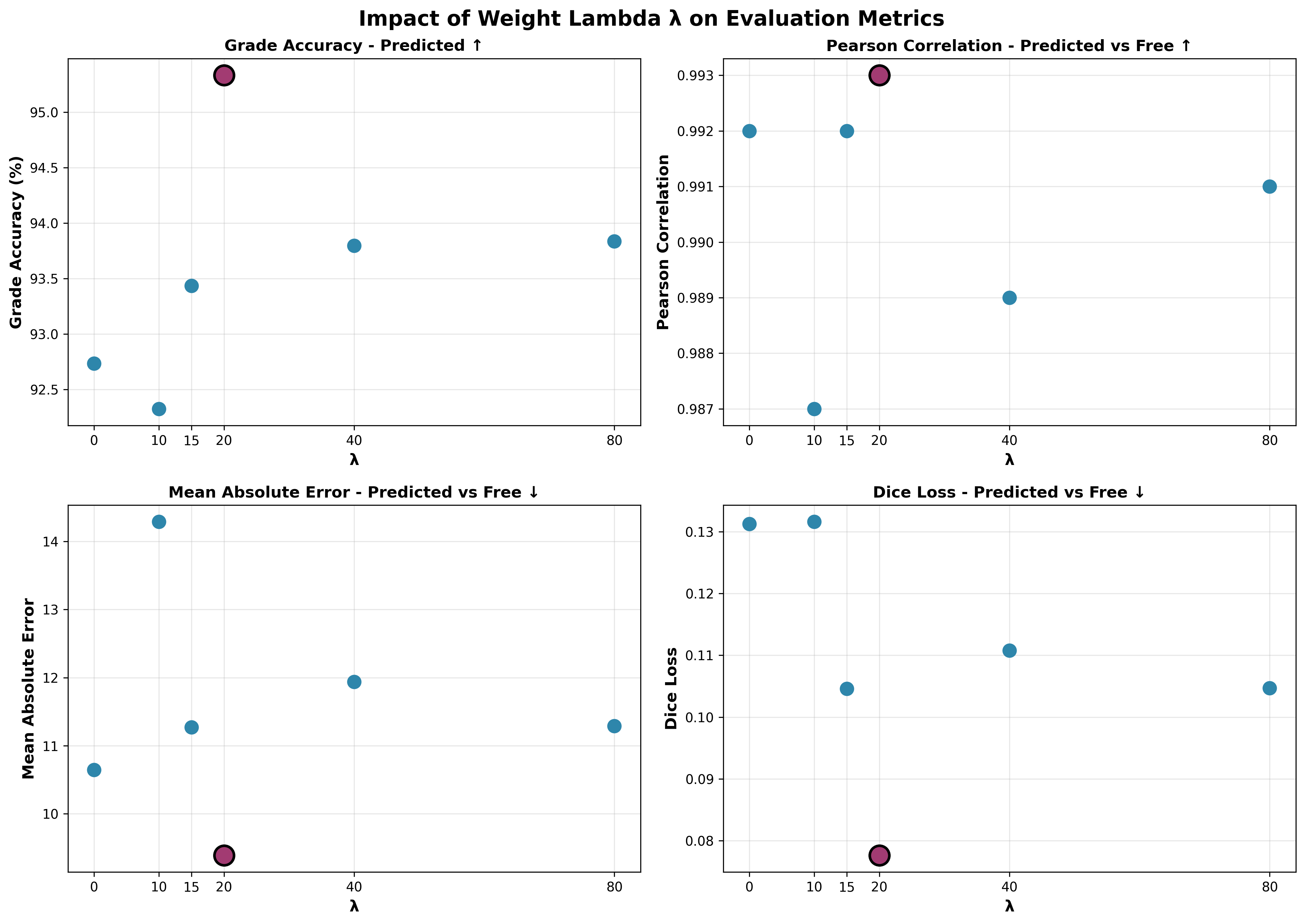}
    \vspace{-5pt}
    \caption{Comparison of different hyperparameter $\lambda$ values used in ProDM and the resulting Agatston score MAE, risk stratification severity grade accuracy, dice loss and Pearson correlation of Agatston scores between predicted output and ground truth. Optimal $\lambda$ is shown in purple.}
    \vspace{-10pt}
    \label{fig:ablative-loss-weight}
\end{figure*}

To assess the impact of weight $\lambda$ for calcium consistency loss on the performance of ProDM, we compare the evaluation metrics Agatston Score MAE,  overall severity grade accuracy, Dice loss, and Pearson correlation for $\lambda=0, 10, 15, 20, 40, 80$ as shown in \textbf{Figure} \ref{fig:ablative-loss-weight}. As $\lambda$ increases from 0, grade accuracy and Dice loss show consistent improvement. In contrast, MAE and Pearson correlation vary across different $\lambda$ values and do not follow a strictly monotonic trend. Notably, $\lambda = 20$ achieves the best overall performance, yielding the highest grade accuracy and Pearson correlation as well as the lowest MAE and Dice loss.

\subsubsection{Impact of Number of Axial Slices}
\begin{figure*}[!t]
    \centering
    \includegraphics[width=0.85\linewidth]{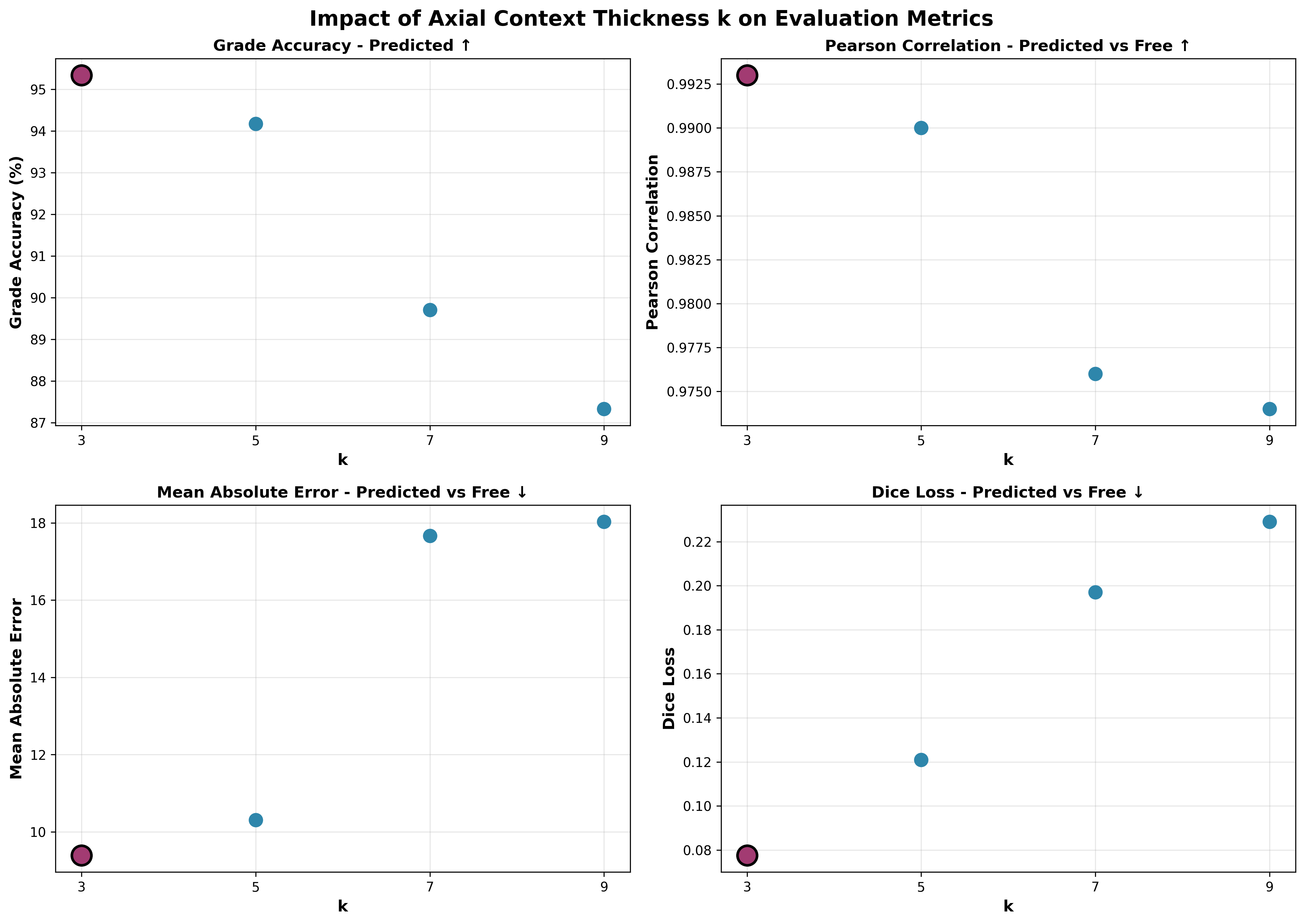}
    \vspace{-5pt}
    \caption{Comparison of different axial context thickness k used in 2.5D training and the resulting Agatston score MAE, risk stratification severity grade accuracy, dice loss, and Pearson correlation of Agatston scores between predicted output and ground truth. Optimal $k$ is shown in purple for each metric.}
    \vspace{-10pt}
    \label{fig:ablative-context-thickness}
\end{figure*}

To investigate the effect of varying the number of slices in the axial direction, we compare the same aforementioned evaluation metrics using different axial context thickness settings $k=3, 5, 7, 9$. As shown in Figure \ref{fig:ablative-context-thickness}, as the number of context slices increased, the performance decreases consistently, where $k=3$ achieves the best performance across all metrics.  

\subsection{Clinical Reader Study}
\label{sec:reader-study}
\begin{figure*}[htbp!]
    \centering
    \includegraphics[width=0.8\linewidth]{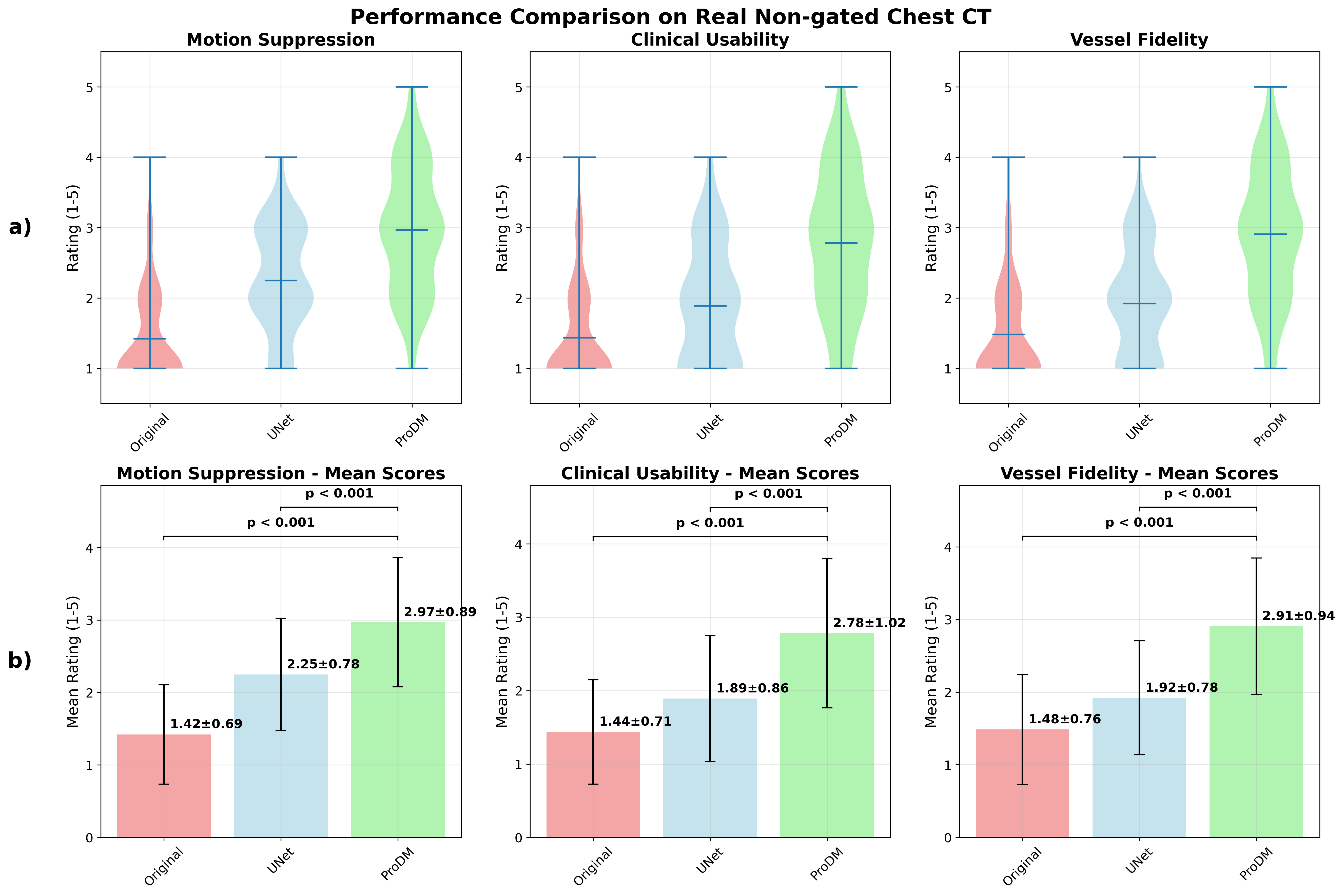}
    \vspace{-5pt}
    \caption{Radiologist assessment results from the reader study. (a) Violin plots show the distribution of radiologists' ratings for motion suppression, clinical usability, and vessel fidelity across methods. (b) Box-and-whisker plots summarize the mean ratings and variability for the original (non-corrected), UNet baseline, and ProDM, with corresponding $p$-values indicating statistical significance.}
    \vspace{-10pt}
    \label{fig:reader-study}
\end{figure*}

With the goal of evaluating our proposed model on real data, we conduct a blinded reader study with two experienced radiologists to assess the performance of baseline (original), UNet, and ProDM on real non-gated scans. The radiologists' preferences and ratings across different evaluation metrics for each sample were collected for further analysis. Figure~\ref{fig:reader-study} (a) shows the distribution of ratings using violin plots, while Figure~\ref{fig:reader-study} (b) summarizes the mean and standard deviation of the ratings for each method across the three evaluation criteria: motion suppression, clinical usability, and vessel fidelity. 

Without motion correction, the original uncorrected samples showed low scores across all three criteria, with motion suppression level $1.42\pm0.69$, clinical usability $1.44\pm0.71$, and vessel fidelity $1.48\pm 0.76$. These scores correspond to moderate to severe motion artifacts, poor clinical usability, and loss of most vessel fidelity, respectively. The violin plot also shows that most scores are centred on the lower end of the Likert scale. 

With UNet correction, all three metrics showed noticeable improvement, with moderate motion suppression ($2.25\pm 0.78)$, limited clinical usability ($1.89\pm 0.86$), and moderate distortions with some loss of vessel fidelity ($1.92\pm0.78$). While all metrics showed some improvements, they are still relatively low and indicate insufficient correction. The violin plots show that the radiologists most frequently gave UNet-corrected outputs a score of around 2.  

ProDM achieved the highest average scores across all criteria, with substantial gains over both the Original and UNet outputs. Motion suppression improved to $2.97 \pm 0.89$, clinical usability reached $2.78 \pm 1.02$, and vessel fidelity increased to $2.91 \pm 0.94$. These indicate that the outputs had some motion artifacts but are mostly clean, good clinical usability with minor limitations, and most vessel details preserved. All improvements of ProDM over the other two methods were statistically significant ($p < 0.001$). The violin plot shows that most scores centered around 3. 

We noticed that uncertainty in mean scores increased across original, UNet, and ProDM; with ProDM having relatively more uncertainties, which is generally expected with improved average performance. However, despite the larger score variance, both radiologists consistently ranked ProDM as the best-performing method when asked to provide an overall ranking of the three approaches in the blinded reader study. 

\section{Discussion}

In this study, we propose a progressive, property-aware motion correction framework \textbf{ProDM} for learning CAC motion correction directly from realistically simulated non-gated CTs synthesized from gated calcium CTs, addressing the critical challenges with a lack of paired data and preserving calcified lesion properties. We evaluate the model on both synthetic and real non-gated data, through extensive quantitative and qualitative analyses. Across all experiments, ProDM demonstrated clear advantages over both conventional UNet-based image-to-image translational baselines and existing diffusion-based approaches. The model achieved substantially lower Agatston score MAE, higher overall and per-class severity grade accuracy, stronger agreement with ground-truth Agatston scores as reflected by a higher Pearson correlation, and lower Dice loss compared with UNet, cGAN, DDPM, and BBDM. This indicates a more faithful recovery of calcified lesion intensity and morphology, as well as better preservation of risk stratification. These improvements were especially prominent in the minimal and mild CAC ranges, where the motion artifacts are more subtle and low intensity, making motion correction especially difficult and resulting in oversmoothing and lesion distortion for traditional methods. 
The ablation studies demonstrated the importance of calcium consistency loss in stabilizing the reconstructed intensity and shape of calcified lesions, as its inclusion results in improvement of precision, recall, and F1 scores across different severity categories. 
Our analysis of the weight hyperparameter $\lambda$ for calcium consistency loss showed that moderate weighting ($\lambda=20$) produced the best overall balance between CAC score preservation and spatial distribution. Similarly, experiments on axial context thickness revealed that a compact 2.5D context window of $k=3$ yields the best performance, whereas more context axial context actually decreases the performance. This may be because calcified lesion motions are typically constrained to only a small number of slices in the axial direction, and larger windows tended to incorporate irrelevant structures, negatively impacting the motion correction ability at inference time. 
Qualitatively, ProDM produced reconstructions that closely resembled motion-free references across diverse motion profiles, yielding the highest SSIM scores with the ground truth.  In the blinded reader study, experienced radiologists consistently ranked ProDM as the most effective method, which achieved good ratings across motion correction, vessel fidelity, and clinical usability, further highlighting our method's potential to be applied on real non-gated CTs for providing opportunistic screening for subclinical atherosclerosis. 

Despite the strong performance of ProDM on both synthetic and real test data, our work is not without limitations. Firstly, there are no ground truth motion-free counterparts available for the real non-gated samples used in the reader study. As a result, we were unable to quantitatively assess objective performance metrics such as Agatston score MAE, severity grade accuracy, Dice loss, and Pearson correlation as done with the synthetic non-gated samples. Additionally, although the reader study results demonstrated that applying ProDM to real non-gated data results in statistically significant improvement in motion correction, clinical usability, and vessel fidelity, these scores are averaged around 3, which corresponds to good clinical quality, but with some limitations, such as remaining motion artifacts. Improving ProDM to consistently achieve ratings of 4 or 5 may require more extensive and realistic simulation pipelines, domain-adaptation strategies, as well as a larger and more diverse reader study to better characterize performance. Another limitation lies in the extensive training time required for a diffusion-based progressive motion correction model. In this work, we used $T=1000$ steps to achieve high-quality and stable motion correction. Although this incurs significant training time, recent work such as \cite{ZHOU2024Shortcut} demonstrates that diffusion-model acceleration techniques can significantly reduce the number of required timesteps. Integrating such methods into ProDM is a promising direction for improving training efficiency. On a related note, training diffusion models directly on 3D medical imaging data is computationally intensive, which motivated our use of a 2.5D training strategy to maintain feasible memory and runtime constraints. This challenge is well recognized, and several recent works have adopted similar 2.5D paradigms for 3D medical imaging data as a practical compromise between computational cost and representational richness \citep{Hu2023DiffGEPCI3M, Chen202425DMA, Jin2025Glioma2p5D, XIE2025103729}. While there may be advantages of extending ProDM to fully 3D, our ablation studies on axial context thickness suggest that performance degrade with including more slices, potentially due to calcium motion being confined to a narrow axial region, and larger context introduces distracting structures that decreases the signal magnitude. Whether a carefully designed 3D variant could overcome these limitations remains an open question.

\section{Conclusion}

We present a novel progressive motion-correction framework for coronary artery calcium in non-gated CTs, which effectively reduces motion artifacts while preserving calcified lesion properties. By simulating non-gated scans with diverse motion profiles directly from gated calcium CT, our approach bypasses the need for paired datasets and allows the model to learn to correct realistic motion artifact patterns. By combining a progressive diffusion framework with calcium consistency loss as guidance, our approach is able to achieve superior performance in reducing motion while effectively preserving the spatial distribution, morphology, and quantity of coronary artery calcium for reliable risk stratification. Together, this formulation enables the potential for opportunistic subclinical atherosclerosis screening and more accurate risk assessment using widely available non-gated CTs.




\bibliographystyle{model2-names.bst}\biboptions{authoryear}
\bibliography{refs}
\clearpage
\appendix
\onecolumn
\section{Additional Implementation Details}
\label{Appendix-A}

\begin{table*}[h]
\centering
\caption{High-level motion profiles used to simulate coronary motion. 
Each trajectory is defined in 3D with axis-specific amplitudes derived from sampled displacement ratios.}
\label{tab:motion_profiles}
\begin{tabular}{l p{4.5cm} p{8cm}}
\toprule
Profile & Behavior & Displacement Model \\
\midrule

Translation
& Slow monotonic drift along a fixed 3D direction 
& $\Vec{d(t)}= (V_x t,V_y t,V_z t)$ \\[0.3em]

Oscillation 
& Back-and-forth motion along a fixed 3D direction 
& $\Vec{d(t)}=\cos(2\pi t + \beta) \cdot (A_x, A_y, A_z)$ where $\beta$ is the phase shift\\[0.3em]

Local Jitter 
& Small, smooth random motion 
& $\Vec{d(t)} = \alpha \sum_{k=1}^{3} w_k \sin(2\pi k t + \beta)\,\vec{u_k}$, where $w_k$ are scaler weights of the $k^{th}$ sinusoidal function, $\vec{u_k}$ is a randomly sampled direction in 3D, $\beta$ is the phase shift, and $\alpha$ is scale applied so $\|\Vec{d(t)}\|\le A$ \\[0.3em]

Piecewise
& Movement in one direction, stops and dwell for some time, then movement in a new direction 
& $d(t)$ follows segment$_1(t)$ with amplitude $A^{(1)}$ in direction $\Vec{u^{(1)}}$,
pauses briefly, then continues along segment$_2(t)$ with amplitude $A^{(2)}$ in direction $\Vec{u^{(2)}}$
\\[0.3em]

\bottomrule
\end{tabular}

\begin{flushleft}
\vspace{0.6em}
\small
\textbf{Note 1.}
Axis-specific amplitudes $(A_x,A_y,A_z)$ were derived from displacement ratios $(r_x,r_y,r_z)$.
For translation, the corresponding velocities are
$V_x = \frac{A_x}{T},\quad V_y = \frac{A_y}{T},\quad V_z = \frac{A_z}{T}$,
where $T$ is the total number of timesteps.\\
\textbf{Note 2.}
We selected the following parameters for the simulations:
$A\in[4,15]$ pixels, $\beta\sim\mathcal{U}(0,2\pi)$, and ratios sampled from
$r_x,r_y\!\sim\![2.0,4.0]$ or $[0.5,1.5]$ depending on the dominant in-plane axis,
and $r_z\!\sim\![0.3,1.0]$ for reduced axial motion.
Piecewise dwell breakpoint time steps, dwell duration, as well as the directions of motion are sampled randomly.
\end{flushleft}
\end{table*}

\begin{table*}[h]
\centering
\caption{Likert-scale scoring criteria used for the reader study. All ratings were provided on a 1-5 scale, where higher scores indicate better performance.}
\begin{tabular}{lll}
\toprule
\textbf{Criterion} & \textbf{Score} & \textbf{Definition} \\
\midrule
\multirow{5}{*}{Motion suppression} 
  & 1 & Severe motion artifacts \\ 
  & 2 & Moderate motion artifacts \\ 
  & 3 & Some motion artifacts remain but mostly clean \\ 
  & 4 & Minor motion artifacts \\ 
  & 5 & No visible motion artifacts \\ 
\midrule
\multirow{5}{*}{Clinical usability} 
  & 1 & Not suitable for clinical use with major artifacts or distortions \\ 
  & 2 & Limited clinical usability with significant quality issues \\ 
  & 3 & Good for clinical use with minor limitations \\ 
  & 4 & High clinical usability with only minor quality issues \\ 
  & 5 & Optimal for clinical use with no quality concerns \\ 
\midrule
\multirow{5}{*}{Vessel fidelity} 
  & 1 & Severe distortion or loss of vessel structures and calcium \\ 
  & 2 & Moderate distortion wth some vessel detail lost \\ 
  & 3 & Minor distortion, most vessel detail preserved \\ 
  & 4 & Minimal distortion with excellent vessel detail preservation\\ 
  & 5 & Perfect preservation of vessel structures and calcium \\ 
\bottomrule
\end{tabular}
\label{tab:likert_scale}
\end{table*}
\clearpage

\section{Additional Results}
\begin{figure*}[h]
    \centering
    \includegraphics[width=0.75\linewidth]{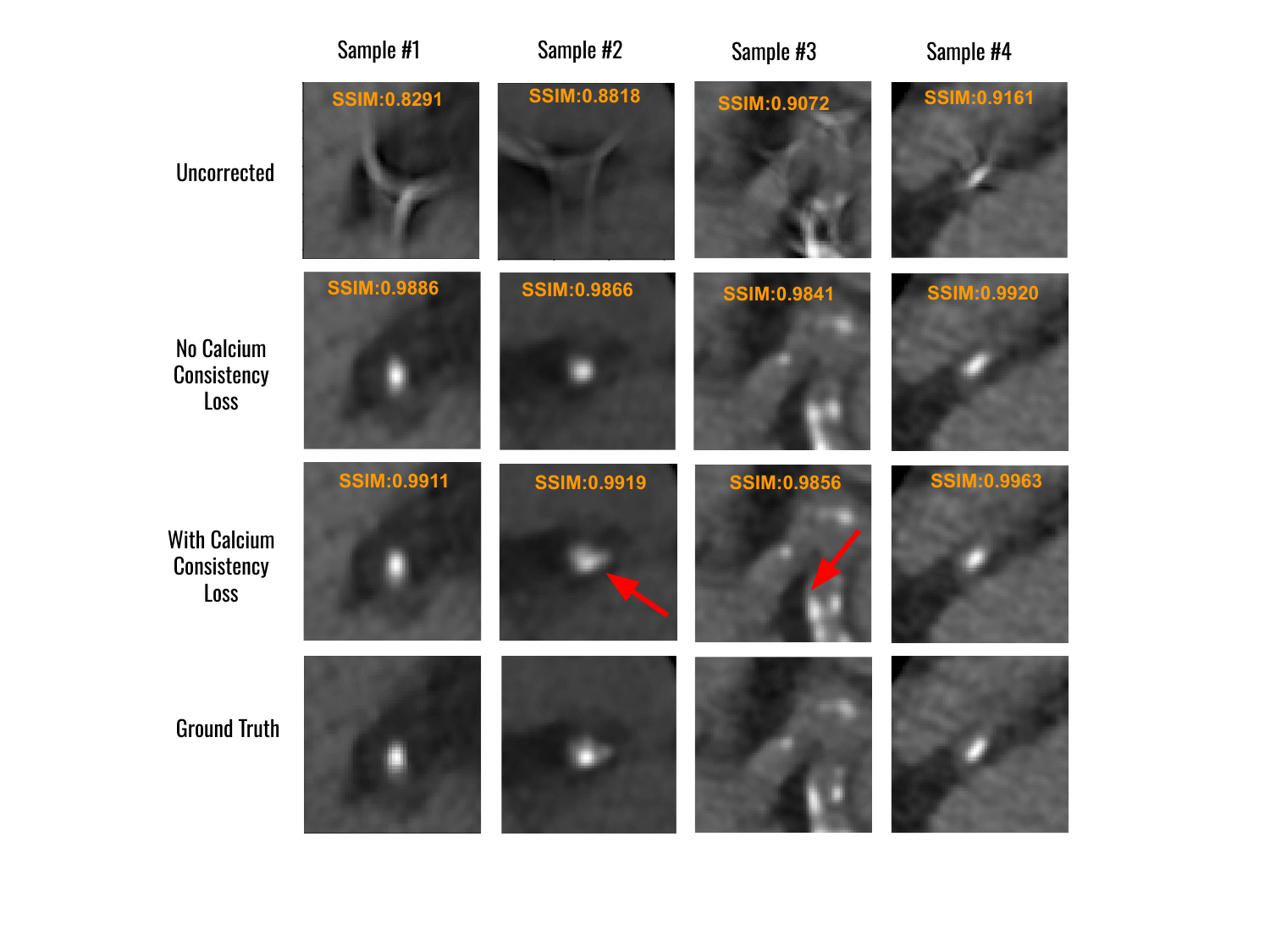}
    \vspace{-5pt}
    \caption{Qualitative comparison of motion correction output of ProDM without calcium consistency loss vs. with calcium consistency loss, compared against ground truth. SSIM against ground truth presented for reference. }
    \vspace{-10pt}
    \label{fig:ablative-calc-consistency-loss}
\end{figure*}
\end{document}